\documentclass[journal,onecolumn]{IEEEtran}
\ifCLASSINFOpdf
\else
\fi
\usepackage{graphicx}

\begin{document}
%
\title{Model Complexity-Accuracy Trade-off for a Convolutional Neural Network}
%
%
%
\author{Atul Dhingra}

\maketitle

\begin{abstract}
Convolutional Neural Networks(CNN) has had a great success in the recent past, because of the advent of faster GPUs and memory access. CNNs are really powerful as they learn the features from data in layers such that they exhibit the structure of the V-1 features of the human brain. A huge bottleneck in this case is that CNNs are very large and have a very high memory footprint, and hence they cannot be employed on devices with limited storage such as mobile phone, IoT etc. In this work we study the model complexity versus accuracy trade-off on MNSIT dataset, and give a concrete framework of handling such a problem, given the worst case accuracy that a system can tolerate. In our work we reduce the model complexity by 236 times, and memory footprint by 19.5 times compared to base model, while achieving worst case accuracy threshold.
\end{abstract}

\begin{IEEEkeywords}
CNN, MNIST, Speed-accuracy Tradeoff, Tensorflow
\end{IEEEkeywords}

%

\section*{Experimental Setup}
The experiment on speed-accuracy trade-off for digit recognition using MNIST dataset \cite{MNIST} is done using tensorflow \cite{TF}. The data is split into training set that contains $55,000$ data points, testing set containing $10,000$ data points and validation set containing $5,000$ data points. Each image in the dataset is a grayscale image of size $28X28$. The labels are "one-hot vectors", which a very sparse vector that contains $0$ in all dimensions except one dimension where it is $1$ corresponding to the true identity. In our work we use the sofmax regression model to compute the loss between the predicted and the ground truth class. The model is trained using Adam optimizer with learning rate $1e-4$ , and mini-batch size of 50 for a total of 20,000 iterations. These parameters are fixed, as we are studying the trade-off of the model complexity and testing accuracy, and these two parameters only help with the convergence time, which we are not considering in our experiment. The convolutional layers weights  and biases are randomly initialized with mean zero($\mu=0$) and standard deviation 0.1($\sigma =0.1$). All the experiments are performed using tensorflow\cite{TF}, using a NVIDIA GeForce GTX Titan X.
\section{Baseline Model}
The baseline model consists of two convolutional layers(conv1,conv2) and two fully connected layers(fc1,fc2) with max pooling layer(pool1,pool2) after conv1 and conv2. There is a dropout layer betwen fc1 and fc2 as shown in the table \ref{table:baseline}
\begin{table}[h!]
\centering
\caption{Baseline Network}
\label{table:baseline}
\begin{tabular}{|l|l|l|l|l|l|}
\hline
\textbf{Name}  & \textbf{Type}   & \textbf{Filter} & \textbf{Output Size} & \textbf{Memory}                                            & \textbf{\#Params}                                                  \\ \hline
\textit{Input} & Image           &                 & 28x28x1              & \begin{tabular}[c]{@{}l@{}}28*28*1\\ =0.784K\end{tabular}  & 0                                                                  \\ \hline
\textit{conv1} & Convolution     & 5x5x1           & 28x28x32             & \begin{tabular}[c]{@{}l@{}}28*28*32\\ =25K\end{tabular}    & \begin{tabular}[c]{@{}l@{}}(5*5*1)*32\\ =800\end{tabular}          \\ \hline
\textit{pool1} & Max Pooling     & 2x2             & 14x14x32             & \begin{tabular}[c]{@{}l@{}}14*14*32\\ =6.272K\end{tabular} & 0                                                                  \\ \hline
\textit{conv2} & Convolution     & 5x5x32          & 14x14x64             & \begin{tabular}[c]{@{}l@{}}14*14*64\\ =12.5K\end{tabular}  & \begin{tabular}[c]{@{}l@{}}(5*5*32)*64\\ =51,200\end{tabular}      \\ \hline
\textit{pool2} & Max Pooling     & 2x2             & 7x7x64               & \begin{tabular}[c]{@{}l@{}}7*7*64\\ =3.1K\end{tabular}     & 0                                                                  \\ \hline
\textit{fc1}   & Fully Connected &                 & 1024                 & 1024                                                       & \begin{tabular}[c]{@{}l@{}}(7*7*64)*1024\\ =3,211,264\end{tabular} \\ \hline
\textit{fc2}   & Fully Connected &                 & 10                   & 10                                                         & \begin{tabular}[c]{@{}l@{}}1024*10\\ =10,240\end{tabular}          \\ \hline
\textit{total} &                 &                 &                      & 48.68K                                                     & 3.27M                                                              \\ \hline
\end{tabular}
\end{table}

The memory footprint of this network is approximately 50K and the number of parameters is 3.27M. The testing accuracy achieved on the baseline model is 99.29\%, and a run time of 165 seconds on Titan X. The As we know the computation speed of convolutional layer is slow, but its memory footprint is small, as evident from table \ref{table:baseline}, on the other hand, the fully connected layers have a high memory footprint, but computationally, it is faster. Keeping these facts in mind, we improved the existing network in the following section.

\section{Optimized Network}
\subsection*{Design Framework}
As in table \ref{table:baseline}, the highest memory contribution is by conv1 and the highest parameter contribution is by fc1. This suggests that to reduce the memory footprint of the network, we need to tweak the conv layers. As the worst-case baseline test accuracy in our case is 95\%, we removed conv2, to see if we can get away with just one convolutional layer, as seen in table \ref{table:-c2}
\begin{table}[h!]
\centering
\caption{Optimization Iteration 1}
\label{table:-c2}
\begin{tabular}{|l|l|l|l|l|l|}
\hline
\textbf{Name}  & \textbf{Type}   & \textbf{Filter} & \textbf{Output Size} & \textbf{Memory}                                            & \textbf{\#Params}                                                  \\ \hline
\textit{Input} & Image           &                 & 28x28x1              & \begin{tabular}[c]{@{}l@{}}28*28*1\\ =0.784K\end{tabular}  & 0                                                                  \\ \hline
\textit{conv1} & Convolution     & 5x5x1           & 28x28x32             & \begin{tabular}[c]{@{}l@{}}28*28*32\\ =25K\end{tabular}    & \begin{tabular}[c]{@{}l@{}}(5*5*1)*32\\ =800\end{tabular}          \\ \hline
\textit{pool1} & Max Pooling     & 2x2             & 14x14x32             & \begin{tabular}[c]{@{}l@{}}14*14*32\\ =6.272K\end{tabular} & 0                                                                  \\ \hline
\textit{fc1}   & Fully Connected &                 & 1024                 & 1024                                                       & \begin{tabular}[c]{@{}l@{}}(14*14*32)*1024\\ =6,422,528\end{tabular} \\ \hline
\textit{fc2}   & Fully Connected &                 & 10                   & 10                                                         & \begin{tabular}[c]{@{}l@{}}1024*10\\ =10,240\end{tabular}          \\ \hline
\end{tabular}
\end{table}
In doing so, we managed to decrease the memory footprint by 15K, but the parameters doubled for fc1 to approximately 6.4M. This network gave a test accuracy of 99.27\% and ran for 164 seconds on Titan X. To reduce the number of parameters, we varied the output size of the fc1 layer as shown in table \ref{table:fc_size}

\begin{table}[h!]
\centering
\caption{Varying filter size of fully connected layer}
\label{table:fc_size}
\begin{tabular}{|l|l|l|}
\hline
\textbf{fc1(size)} & \textbf{Test Accuracy} & \textbf{Time(Titan X)} \\ \hline
\textit{1024}      & 99.27                  & 164 s                  \\ \hline
\textit{512}       & 99.25                  & 155 s                  \\ \hline
\textit{256}       & 99.21                  & 148 s                  \\ \hline
\textit{\textbf{128}}       & 99.1                   & 145 s                  \\ \hline
\textit{64}        & 98.9                   & 159 s                  \\ \hline
\textit{32}        & 98.7                   & 158 s                  \\ \hline
\end{tabular}
\end{table}
We observe when fc1 is 128 in size, it gives the best trade-off for running time without affecting the accuracy too much. By choosing 128 as fc1 size, we have dramatically decreased the number of parameters in fc1 from 6.4M in table \ref{table:-c2} to 0.8M in this case. This motivated us to use 128 size fc1 in our final model. \\

After all these tweaks, still the memory footprint is high, and the number of parameters in millions. To further reduce the size and parameters, we tweaked the conv1 filter size, and varied it to 3x3 and 1x1. we observed that 3x3 gave us the best tradeoff betwen the number of parameters and the testing accuracy. We also varied the depth of the convolutional filter to see if actually need 32 filters to achieve 95\% accuracy. The results below use the fc1 layer of size 128, while varying the conv1 layer in table \ref{table:5x5},\ref{table:3x3},\ref{table:1x1}

\begin{table}[h!]
\centering
\caption{Varying depth on 5x5 conv1 filter}
\label{table:5x5}
\begin{tabular}{|l|l|l|}
\hline
\textbf{Filter Size} & \textbf{Test Accuracy} & \textbf{Time in seconds(on Titan X)} \\ \hline
\textit{5x5x32}      & 98.6                   & 114                                  \\ \hline
\textit{5x5x16}      & 98.5                   & 110                                  \\ \hline
\textit{5x5x8}       & 98.2                   & 107                                  \\ \hline
\textit{5x5x4}       & 97.8                   & 109                                  \\ \hline
\textit{5x5x2}       & 97.38                  & 108                                  \\ \hline
\end{tabular}
\end{table}

\begin{table}[]
\centering
\caption{Varying depth on 3x3 conv1 filter}
\label{table:3x3}
\begin{tabular}{|l|l|l|}
\hline
\textbf{Filter Size} & \textbf{Test Accuracy} & \textbf{Time in seconds(on Titan X)} \\ \hline
\textit{3x3x32}      & 98                     & 117                                  \\ \hline
\textit{3x3x16}      & 97.79                  & 108                                  \\ \hline
\textit{3x3x8}       & 97.5                   & 103                                  \\ \hline
\textit{3x3x4}       & 96.6                   & 108                                  \\ \hline
\textit{3x3x2}       & 96.2                   & 106                                  \\ \hline
\end{tabular}
\end{table}

\begin{table}[h!]
\centering
\caption{Varying depth on 1x1 conv1 filter}
\label{table:1x1}
\begin{tabular}{|l|l|l|}
\hline
\textbf{Filter Size} & \textbf{Test Accuracy} & \textbf{Time in seconds(on Titan X)} \\ \hline
\textit{1x1x32}      & 95.88                  & 111                                  \\ \hline
\textit{1x1x16}      & 95.9                   & 106                                  \\ \hline
\textit{1x1x8}       & 95.0                   & 105                                  \\ \hline
\textit{1x1x4}       & 94.38                  & 102                                  \\ \hline
\textit{1x1x2}       & 94.28                  & 100                                  \\ \hline
\end{tabular}
\end{table}
From this data we observe that the accuracy dips more rapidly as the conv1 filter size is decreased. For example for 5x5 filter in table \ref{table:5x5} the accuracy dips from 98.6 to 97.38, that is about 1\% decrease whereas in 3x3 filter, the accuract dips from 98 to 96.2, which is almost double. Hence for our design, we stick with the 5x5 filter, as in our final design we use a depth 2 filter, and only 32 parameters are added(50-18), which is minuscule as compared to the order of parameters in thousands.  We also use a 4x4 pooling layer instead of the 2x2 pooling layer in baseline to reduce the parameters from fc1 layer without losing much test accuracy.

\subsection*{Final Model}
We designed the following network as shown in table \ref{table:OPT} as per the worst case threshold of 95\% test accuracy. 
\begin{table}[h!]
\centering
\caption{Optimized Network}
\label{table:OPT}
\begin{tabular}{|l|l|l|l|l|l|}
\hline
\textbf{Name}  & \textbf{Type}   & \textbf{Filter} & \textbf{Output Size} & \textbf{Memory}                                           & \textbf{\#Params}                                          \\ \hline
\textit{Input} & Image           &                 & 28x28x1              & \begin{tabular}[c]{@{}l@{}}28*28*1\\ =0.784K\end{tabular} & 0                                                          \\ \hline
\textit{conv1} & Convolution     & 5x5x2           & 28x28x2              & \begin{tabular}[c]{@{}l@{}}28*28*2\\ =1.5K\end{tabular} & \begin{tabular}[c]{@{}l@{}}(5*5*1)*2\\ =50\end{tabular}    \\ \hline
\textit{pool1} & Max Pooling     & 4x4             & 7x7x2                & \begin{tabular}[c]{@{}l@{}}7*7*4\\ =98\end{tabular}   & 0                                                          \\ \hline
\textit{fc1}   & Fully Connected &                 & 128                  & 128                                                       & \begin{tabular}[c]{@{}l@{}}(7*7*2)*128\\ =12.5K\end{tabular} \\ \hline
\textit{fc2}   & Fully Connected &                 & 10                   & 10                                                        & \begin{tabular}[c]{@{}l@{}}128*10\\ =1.2K\end{tabular}     \\ \hline
\textit{total} &                 &                 &                      & 2.5K                                                    & 13.8K                                                      \\ \hline
\end{tabular}
\end{table}
This optimized model produces a test accuracy of 95.81\%. This optimized network has 19.5 times smaller memory footprint as compared to the baseline model. It also requires 236 times less parameter computations as compared to the baseline model as detailed in the table \ref{table:comp}

\begin{table}[h!]
\centering
\caption{Comparison of Baseline Network with Optimized Network}
\label{table:comp}
\begin{tabular}{|l|l|l|}
\hline
\textit{\textbf{}} & \textbf{Baseline} & \textbf{Optimized} \\ \hline
\textit{\#Params}  & 3.27M             & 13.8K              \\ \hline
\textit{Memory}    & 48.65K            & 2.5K              \\ \hline
\textit{Run Time(Titan X)}  & 165sec            & 98sec              \\ \hline
\end{tabular}
\end{table}

Note: With the same architecture, but using 3x3 conv1 instead of 5x5, although we get a reduction of 32 parameters(5*5*2-3*3*2), but the test accuracy goes to 93.66\% which violates our design parameters.

\section{Model Size and Accuracy Comparison}
\begin{figure}[h]
\begin{center}

\includegraphics[scale=0.8]{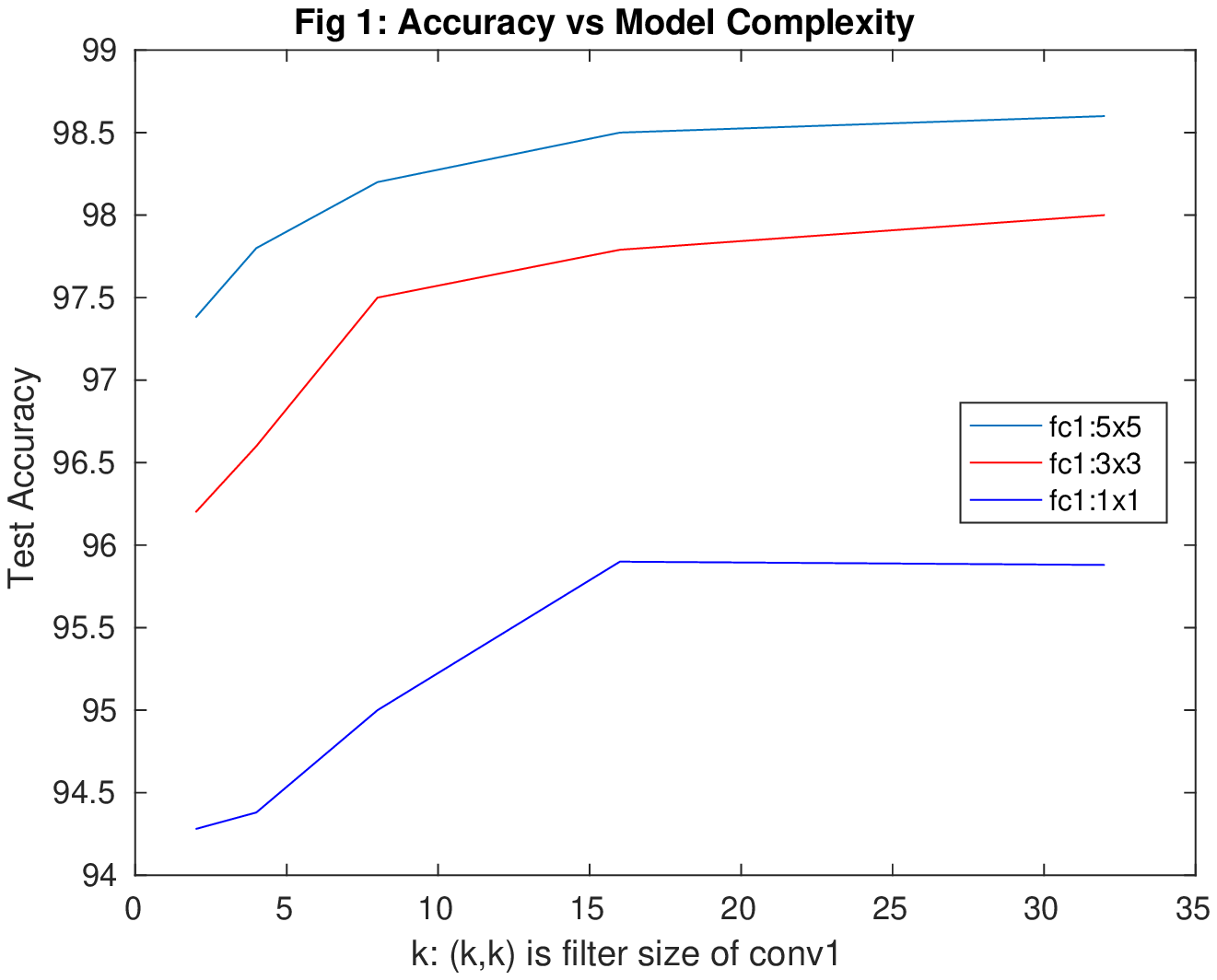}

\label{fig:comp}
\end{center}
\end{figure}

Given $s(f)$ is the minimum model size that can achieve the accuracy f, we can assert that 
\begin{equation}
\label{eq:Q3}
0 < s(f_1) < s(f_2)  ... <s(f_n) < 1
\end{equation} 
such that, $0 \leq f_1 < f_2 < ... < f_n  \leq 1$
This is evident from fig  ,as the model size decreases, the testing accuracy decreases as shown in fig \ref{fig:comp}

Take a slice of fig \ref{fig:comp} at k=32, at this point the 5x5 curve is above 3x3 , which is above 1x1 curve. at this slice k=32, the complexity of the 5x5 network is greater than 3x3, which is greater than 1x1. Similarly the test accuracy at this point follows the same trend. What we can assert is that, as the other parameters of the model is same, and we decrease the filter size from 5x5 to 3x3 followed by 1x1, the model complexity of the network is decreasing. But, as we see from the fig \ref{fig:comp}, the test accuracy also follows the same trend, therefore we can assert that the claim made in equation \ref{eq:Q3} is valid. This can also be realized by the fact that the Optimal network proposed in this scenario has a 5x5 filter, which has a test accuracy of 95.81\%, but if instead we use a 3x3 network(details committed), which is a much less complex model, the accuracy percentage drops to 93.66\% which satisfies claim made in equation \ref{eq:Q3}. As all the curves are monotonically decreasing in the negative x-axis, having proved the fact for k=32, this claim will still hold of k=16,k=8..etc. 

\section{Conclusion}
Based on our experiments, we can conclude that given a baseline worst case test accuracy, we can systematically reduce the model complexity by following a structured approach and handling the trade-off between memory overhead and parameter required for tuning. For our experiment on MNIST\cite{MNIST} we reduced the parameters by 236 times, and memory footprint by 19.5 times as compared to the base model, and achieving the lower test accuracy threshold of 95\% .

\end{document}